\documentclass[letterpaper, 10 pt, conference]{ieeeconf}  

\IEEEoverridecommandlockouts                              

\usepackage{graphics} 
\usepackage{epsfig} 
\usepackage{amsmath} 
\usepackage{amssymb}  
\usepackage{bm} 
\usepackage{textgreek} 
\usepackage{flushend} 
	
\usepackage{comment}	
\usepackage{siunitx} 
\usepackage[percent]{overpic}
\usepackage{xcolor}
\usepackage{pgfplots}
\pgfplotsset{compat=newest}
\usetikzlibrary{plotmarks}
\usetikzlibrary{arrows.meta}
\usepgfplotslibrary{patchplots}
\usepackage{grffile}
\pgfplotsset{plot coordinates/math parser=false}
\newlength\fwidth
\newlength\fheight
\usepackage{url}
\usepackage[hidelinks]{hyperref}

\newcommand{\normtwosq}[1]{\lVert #1 \rVert_2^2}
\usepackage{cite}
\allowdisplaybreaks

\title{\LARGE \bf
Multi-Modal Decentralized Reinforcement Learning \\ for Modular Reconfigurable Lunar Robots}

\author{Ashutosh Mishra$^{1}$, Shreya Santra$^{1}$, Elian Neppel$^{1}$, Edoardo M. Rossi Lombardi$^{1,2}$,\\ Shamistan Karimov$^{1}$, Kentaro Uno$^{1}$, and Kazuya Yoshida $^{1}$
\thanks{$^{*}$This work was supported by JST Moonshot R\&D Program, Grant Number JPMJMS223B.}
\thanks{
$^{1}$A. Mishra, S. Santra, E. Neppel, S. Karimov, K. Uno,  and K. Yoshida are with the Space Robotics Lab. (SRL) in Department of Aerospace Engineering, Graduate School of Engineering, Tohoku University, Sendai 980--8579, Japan.  
    }%
\thanks{
$^{2}$E. M. R. Lombardi is with the Politecnico di Milano, Italy.
    }
\thanks{
\textit{The corresponding author is Ashutosh Mishra (E-mail: \tt{ashutosh.mishra@dc.tohoku.ac.jp}).}
    }%
}%

\begin{document}
\bstctlcite{refs:BSTcontrol}

\maketitle
\thispagestyle{empty}
\pagestyle{empty}

\begin{abstract}
Modular reconfigurable robots suit task-specific space operations, but the combinatorial growth of morphologies hinders unified control. We propose a decentralized reinforcement learning (Dec-RL) scheme where each module learns its own policy: wheel modules use Soft Actor-Critic (SAC) for locomotion and 7-DoF limbs use Proximal Policy Optimization (PPO) for steering and manipulation, enabling zero-shot generalization to unseen configurations. In simulation, the steering policy achieved a mean absolute error of \SI{3.63}{\degree} between desired and induced angles; the manipulation policy plateaued at \SI{84.6}{\percent} success on a target-offset criterion; and the wheel policy cut average motor torque by \SI{95.4}{\percent} relative to baseline while maintaining \SI{99.6}{\percent} success. Lunar-analogue field tests validated zero-shot integration for autonomous locomotion, steering, and preliminary alignment for reconfiguration. The system transitioned smoothly among synchronous, parallel, and sequential modes for Policy Execution, without idle states or control conflicts, indicating a scalable, reusable, and robust approach for modular lunar robots.
\end{abstract}
\section{Introduction}

Modular robotics presents an exciting frontier for space exploration, where mission success depends on reliability, ease of repair, and adaptability. In line with the Japan Science and Technology Agency's Moonshot Project (Goal 3)~\cite{moonshot}, our long-term aim is to develop modular robots that autonomously learn, adapt, evolve in intelligence, and collaborate with humans on the lunar surface by 2050. To achieve this goal, we developed “MoonBots,” a modular and reconfigurable platform designed for autonomous and cooperative operations in assembling and maintaining future lunar infrastructure~\cite{moonbot}.

While, teleoperation and algorithmic assembly show promise~\cite{ashutosh_2025, luca, camille}, autonomous self-reconfiguration is essential for long missions. Modular robots combine standard units into task-specific morphologies. Although hardware advances allow rapid reconfiguration, learning control policies remains challenging when varied combinations alter kinematics and observation–action spaces~\cite{iwata_adaptive_2021, whitman_learning_2023}.

\begin{figure}[t]
    \centering
    \includegraphics[width=0.9\linewidth]{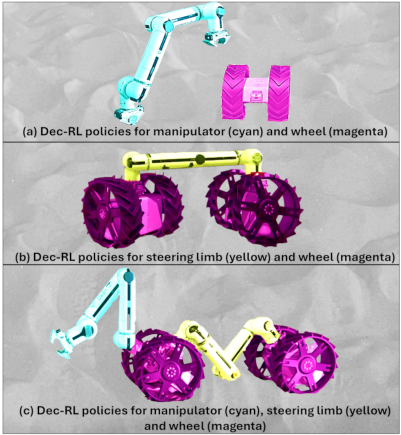}
    \caption{MoonBot modules and policy: Wheel modules use SAC (Soft Actor-Critic) for locomotion and 7-DoF limbs use PPO (Proximal Policy Optimization) for manipulation and steering.}
    \label{fig:figure1}
\end{figure}

As shown in Fig.~\ref{fig:figure1}, MoonBots consist of interchangeable modules capable of reconfiguration to transport, assembly, and manipulation. However, learning separate policies for each configuration is inefficient, while a single monolithic policy struggles to generalize due to conflicting optimization signals and lack of structural inductive bias~\cite{whitman_learning_2023, sartoretti_distributed_2019}. A promising solution is decentralized reinforcement learning (Dec-RL), where each module learns its own policy from local observations while coordinating with neighbors to achieve global goals~\cite{wang_distributed_2022, leottau_decentralized_2017, busoniu_decentralized_2006}. This work proposes a decentralized architecture for modular, reconfigurable robots, allowing for reusable and plug-and-play policies that generalize to new morphologies without requiring retraining, thereby enabling rapid adaptation to new hardware layouts in unpredictable lunar environments.

The key contributions of this work are:
\begin{enumerate}
    \item A fully decentralized RL architecture where each module learns independently with task-specific algorithms, enabling scalable control and zero-shot adaptation to new configurations without retraining.
    \item Enhanced sample efficiency and robustness over centralized approaches, maintaining performance across varied configurations.
    \item Experimental validation with experiments conducted in lab and lunar-like Space Exploration Field at JAXA \cite{jaxafield}, demonstrating autonomous reconfiguration and complex coordinated task execution.
\end{enumerate}

\section{Related Work}

Lunar operations are challenged by high-latency, low-bandwidth communications and unpredictable environments. 
Advances in AI and robotics are enabling greater autonomy and reducing reliance on Earth-based teleoperation. Deep Reinforcement Learning (DRL) enables adaptive autonomy in space robots by learning optimal actions in stochastic, partially observable environments typical of planetary surfaces~\cite{wang_distributed_2022}. Recent efforts have also aimed to establish standardized evaluation frameworks for robotic manipulation in space, highlighting the importance of reproducible benchmarks for comparing learning-based and classical controllers~\cite{OrsulaTowardsBR}. Unlike rule-based algorithmic systems, RL controllers generalize across varied and unseen conditions, aiding anomaly recovery, re-planning, and context-aware exploration while reducing human intervention~\cite{whitman_learning_2023, wang_model-based_2021, han_reinforcement_2022}. 

As system complexity grows - as in a multi-modular reconfigurable robotic system - centralized RL becomes impractical, less scalable, and more vulnerable to failures~\cite{busoniu_decentralized_2006, wang_distributed_2022}. Dec-RL aims to overcome these challenges by distributing learning and execution, enabling each module to act independently with partial observability while achieving coordinated goals~\cite{zhang_deep_2023, nair_decentralizing_2022}. Early work showed Dec-RL can outperform centralized baselines in sample and computational efficiency~\cite{busoniu_decentralized_2006, gal_offline_2023}, and recent studies have applied it to multi-robot systems for scalable, robust cooperation~\cite{wang_distributed_2022, zhang_deep_2023}. Examples include Sartoretti et al.~\cite{sartoretti_distributed_2019} using decentralized policies for articulated mobile robots, Zhang et al.~\cite{zhang_decentralized_2020} enabling cooperative object transport via decentralized Deep Q-Learning (DQN) without a central controller, and Farivarnejad et al.~\cite{farivarnejad_fully_2021} implementing a decentralized Proportional–Derivative (PD) controller. for payload transport in microgravity with minimal information sharing. Whitman et al.~\cite{whitman_learning_2023} proposed a policy structure for modular robotics, where policy graphs match hardware graphs, sharing networks for module types to enable transfer to unseen configurations. However, their approach relies on a configuration-conditioned global policy and is not fully decentralized.

Our proposed architecture addresses this gap where locomotion and manipulation modules autonomously acquire reusable policies that can be composed at runtime. We validate our method with simulations and hardware experiments across diverse mission tasks and robot morphologies.

\section{Modular Robot System and Decentralized Policy Learning}
\begin{figure}[t]
    \centering
    \begin{minipage}[t]{0.48\columnwidth}
        \centering
        \begin{overpic}[width=\linewidth]{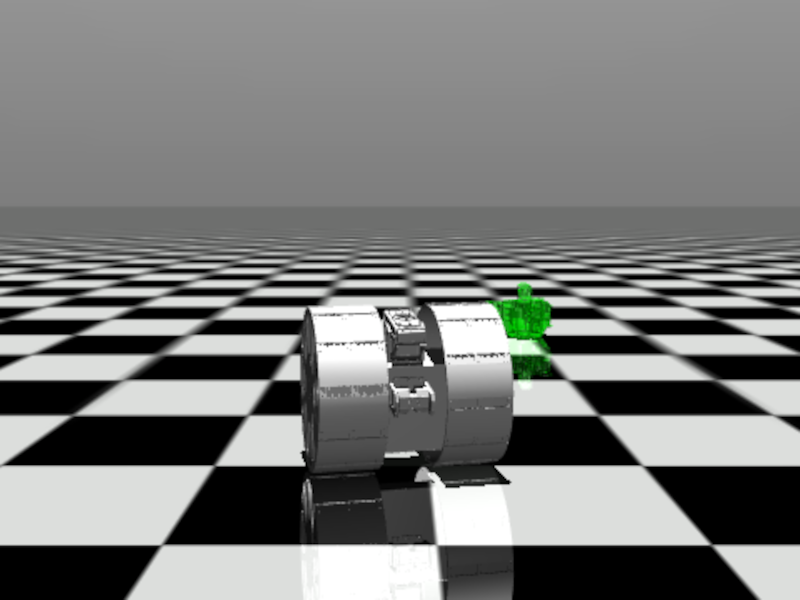}
            \put(3,3){\fcolorbox{black}{black}{\textcolor{white}{\textbf{(a)}}}}
        \end{overpic}
    \end{minipage}%
    \hspace{0.01\columnwidth}%
    \begin{minipage}[t]{0.48\columnwidth}
        \centering
        \begin{overpic}[width=\linewidth]{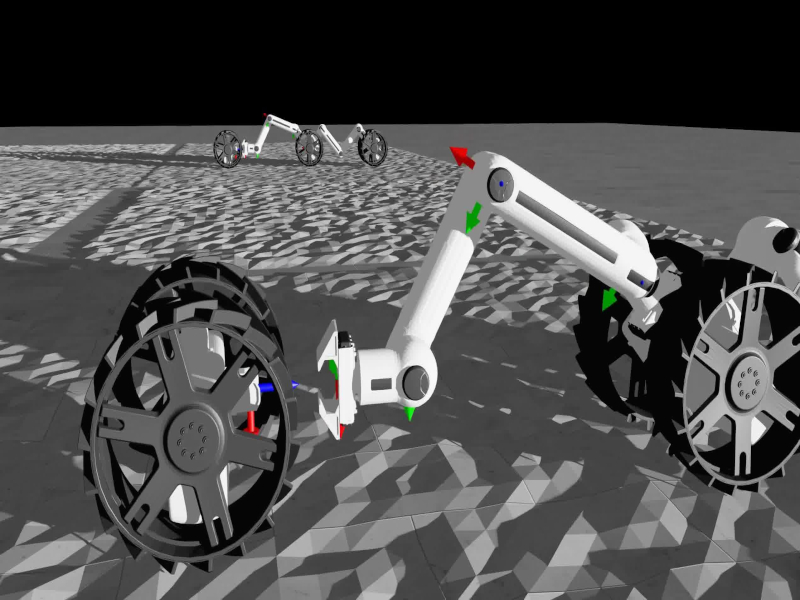}
            \put(3,3){\fcolorbox{black}{black}{\textcolor{white}{\textbf{(b)}}}}
        \end{overpic}
    \end{minipage}

    \par\vspace{4pt} 

    \begin{minipage}[t]{0.48\columnwidth}
        \centering
        \begin{overpic}[width=\linewidth]{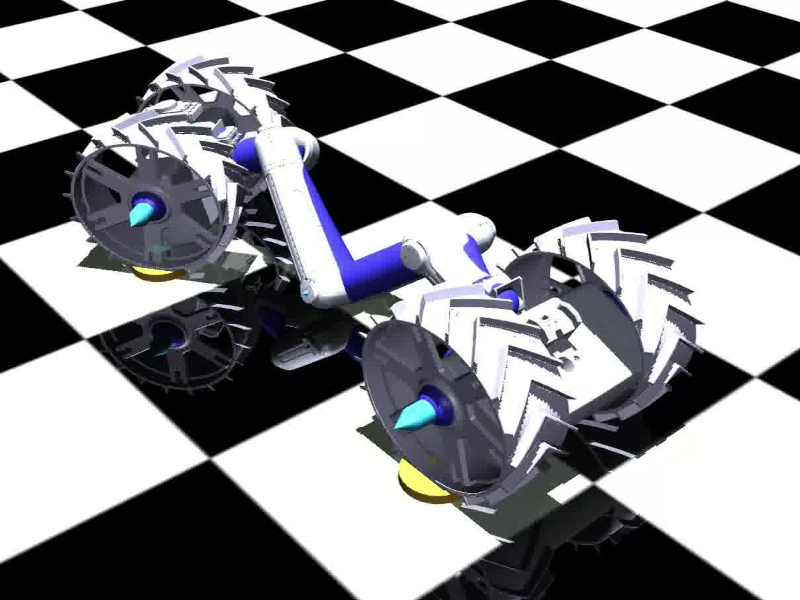}
            \put(3,3){\fcolorbox{black}{black}{\textcolor{white}{\textbf{(c)}}}}
        \end{overpic}
    \end{minipage}%
    \hspace{0.01\columnwidth}%
    \begin{minipage}[t]{0.48\columnwidth}
        \centering
        \begin{overpic}[width=\linewidth]{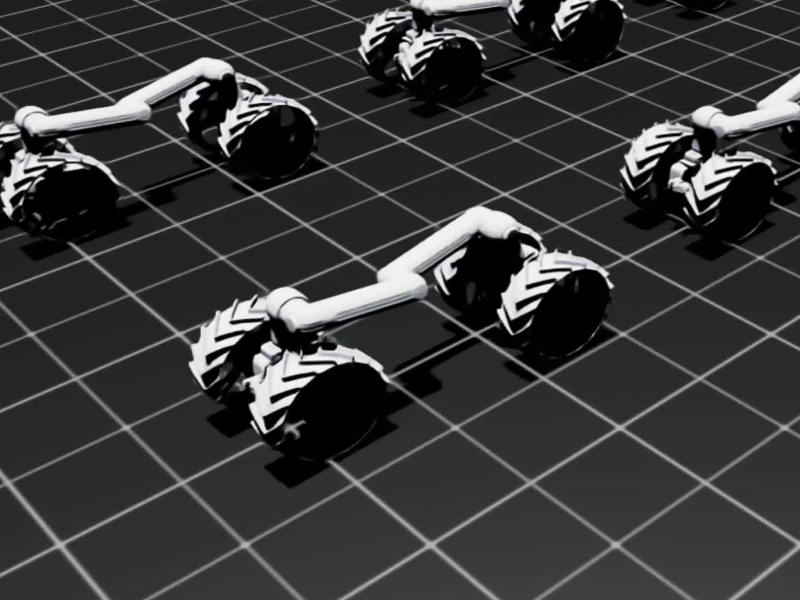}
            \put(3,3){\fcolorbox{black}{black}{\textcolor{white}{\textbf{(d)}}}}
        \end{overpic}
    \end{minipage}

    \caption{
        Simulation environment for modular robot components, used for individual policy learning and testing. Wheel locomotion and Dragon-mode steering were learned in MuJoCo, while manipulation and Vehicle-mode steering tasks were implemented in NVIDIA Isaac Lab. (a) Wheel dynamics simulation. (b) Manipulator grasping. (c) Dragon steering. (d) Vehicle limb-steering.
    }
    \label{fig:simulations}
\end{figure}

The robotic platform comprises primary modular units that integrate a two-motor wheel base with a 7-DoF robotic limbs as shown in Fig. \ref{fig:figure1}(a), supporting flexible morphologies for lunar construction and manipulation. We focus on two configurations: \emph{Vehicle} mode shown in Fig. \ref{fig:figure1}(b), which employs two wheel modules and a single limb for active steering; and \emph{Dragon} mode Fig. \ref{fig:figure1}(c), consisting of two serially connected wheel-limb modules, with a front limb for manipulation and a rear limb for stabilization and maneuvering, enabling concurrent locomotion and task execution. These configurations, along with the corresponding simulation environments used for their policy learning, are shown in Fig.~\ref{fig:simulations}. While the manipulation and locomotion tasks could be solved using several alternative reinforcement learning algorithms, our aim is not to compare specific algorithms, but to demonstrate how heterogeneous policies can coexist and cooperate within a single modular architecture. PPO for manipulation and SAC for locomotion were chosen owing to their proven stability and effectiveness in continuous-control tasks~\cite{schulman_ppo_2017,haarnoja_sac_2018}. This combination enables the evaluation of a multi‑modal decentralized learning framework, where each module type learns its own control strategy yet can be composed dynamically at runtime. The key contribution lies in showing that such independently trained, task‑specific policies can coordinate seamlessly on modular lunar hardware, supporting adaptability rather than algorithm novelty.
Coordinating these heterogeneous modules with multiple degrees of freedom requires efficient control. We leverage the Motion Stack framework by Neppel et al.~\cite{neppel2025robustmodularmultilimbsynchronization}, which enforces smooth multi-limb trajectory synchronization at a low level via state constraints within a hypersphere. The approach dynamically adapts to system variations and disturbances, providing a robot-agnostic, modular interface that is well-suited for our diverse morphologies.
A key difficulty arises from significant variations in robot topology, as a single control policy seldom transfers reliably across morphologies due to prohibitive data requirements, high computational costs, and sensitivity to configuration changes. Policies trained for one configuration (e.g., \emph{Vehicle} mode) may degrade or destabilize when applied to another (e.g., \emph{Dragon} mode) due to differences in dynamics, control dimensionality, and module interactions. To address this, we employ a modular reinforcement learning framework in which policies for each module type—wheel bases, steering limbs, and manipulation limbs—are trained independently and composed at runtime to match the active morphology. Such a paradigm is well-suited for reconfigurable lunar robots, supporting adaptability, autonomy, and resilience.

Agents use episodic learning with a \SI{0.01}{s} control step and a \SI{5}{s} horizon. Episodes terminate on success, instances of safety violations, or exceeding the time limit. Returns use generalized advantage estimation (GAE); values are bootstrapped at non-terminal and time-limit truncations and are not bootstrapped at true terminals. Parameters used for PPO (arm) are Steps-per-env 128; minibatches 64; learning rate $1\times10^{-4}$; $\gamma=0.99$; $\lambda=0.95$; clip $0.2$; entropy coefficient $0.002$; value-loss coefficient $1.0$; target-KL $0.008$, whereas for SAC (wheels) Actor/critic Adam $3\times10^{-4}$; $\gamma=0.99$; target-update $\tau=0.005$; batch 256; replay $10^{6}$; automatic entropy tuning with learned target.

We represent the robot morphology as a graph \(\mathcal{G} = (\mathcal{M}, \mathcal{E})\), where \(\mathcal{M}\) is the set of modules (wheels and limbs) and \(\mathcal{E}\) the set of edges encoding mechanical and control connections. Each module \(m \in \mathcal{M}\) is modeled as a local Markov Decision Process (MDP) \(\mathcal{M}_m = (\mathcal{S}_m, \mathcal{A}_m, \mathcal{T}_m, r_m, \gamma)\), receiving local observations \(o_m\), executing actions \(a_m\), and obtaining rewards \(r_m\). The global system forms a Decentralized Partially Observable MDP (Dec-POMDP), with policies \(\pi_m(a_m|o_m)\) operating based solely on local information and limited synchronization, enabling scalability in constrained or fault-prone settings.
The notation $\|x\|_2$ denotes the Euclidean ($\ell_2$) norm, and $\|x\|_2^2$ its square.
\paragraph{Wheel Module Policy}  
State: \(s_t^{\text{wheel}} = [p_t, \dot{p}_t, \Delta_g]\), where \(p_t\) is position, \(\dot{p}_t\) velocity, \(\Delta_g = p_g - p_t\) relative goal displacement.  
Action: $a_t^{\text{wheel}}=(a_t^{\text{turn}},a_t^{\text{move}})$, where the SAC actor outputs $u_t^{\text{turn}},u_t^{\text{move}}\in[-1,1]$; we set $a_t^{\text{turn}}=u_t^{\text{turn}}$ and map $u_t^{\text{move}}$ to $\{-1,0,1\}$ via a small deadband/hysteretic rule in the environment, optimized via SAC \cite{haarnoja_sac_2018}.  
The reward function for the wheel module is defined as:
\begin{align}
\label{eq:wheel_reward}
r_t^{\text{wheel}} ={}& -\alpha \normtwosq{p_t - p_g} \nonumber \\
                       & - \beta \normtwosq{\tau_t} \nonumber \\
                       & - \lambda \mathbb{I}_{\text{unstable}} + \eta \mathbb{I}_{\text{goal}},
\end{align}
where $\tau_t$ is the motor torque; the indicator functions penalize instability and reward goal attainment; and $\alpha$, $\beta$, $\lambda$, and $\eta$ are weighting coefficients corresponding to position error, torque penalty, instability, and goal reward terms, respectively (Eq.~\ref{eq:wheel_reward}). Learning occurs on procedurally generated terrains with friction \(\mu \in [0.3,1.0]\) for generalization.
\paragraph{Steering Limb Policy}  
State: \(s_t^{\text{steer}} = [q_t, \dot{q}_t, \phi_t]\), joint angles/velocities \(q_t, \dot{q}_t \in \mathbb{R}^7\), desired steering angle \(\phi_t\).  
Action: \(\Delta q_t \in \mathbb{R}^7\), incremental joint position commands.  
Reward:  
\begin{align}
\label{eq:steer_reward}
r_t^{\text{steer}} ={}& - \gamma_1 (\hat{\phi}_t - \phi_t)^2 \nonumber \\
                      & - \gamma_2 \operatorname{OutOfPlane}(q_t) - \gamma_3 \operatorname{JointLimitPenalty}(q_t),
\end{align}
where \(\hat{\phi}_t\) is the curvature induced by deformation of the wheel-limb-wheel axis (Eq.~\ref{eq:steer_reward}), computed as the signed planar angle between the forward-facing unit vectors of the front and rear wheel modules (Eq.~\ref{eq:steer_angle}), obtained from their body quaternions by projecting the local wheel-forward axis onto the ground plane and using the \(\operatorname{atan2}\) function. This encourages compliant planar steering within joint limits.

\begin{align}
\label{eq:steer_angle}
\hat{\phi}_t = \operatorname{atan2}\big( (\mathbf{f}_\text{front} \times \mathbf{f}_\text{rear}) \cdot \mathbf{\hat{z}},\;
\mathbf{f}_\text{front} \cdot \mathbf{f}_\text{rear} \big),
\end{align}
where \(\mathbf{f}_\text{front}\) and \(\mathbf{f}_\text{rear}\) are the normalized forward-facing vectors of the front and rear wheel modules projected onto the horizontal plane, and \(\mathbf{\hat{z}}\) is the upward unit vector.

\paragraph{Manipulation limb Policy}: State: \(s_t^{\text{manip}} = [q_t, \dot{q}_t, x_t^{\text{ee}}, x_t^{\text{goal}}]\), with end-effector pose \(x_t^{\text{ee}} \in \mathbb{R}^3\) and target \(x_t^{\text{goal}} \in \mathbb{R}^3\).  
Action: \(\Delta q_t \in \mathbb{R}^7\).  
Optimized via PPO \cite{schulman_ppo_2017}.  
Reward:
\begin{align}
\label{eq:manip_reward}
r_t^{\text{manip}} ={}& -\delta \normtwosq{x_t^{\text{ee}} - x_t^{\text{goal}}}\nonumber \\
                      & - \epsilon \normtwosq{\Delta q_t} + \zeta \mathbb{I}_{\text{success}},
\end{align}
encouraging spatial accuracy, smooth actuation, and task completion (\ref{eq:manip_reward}).

\begin{figure*}
    \centering
    \includegraphics[width=\textwidth]{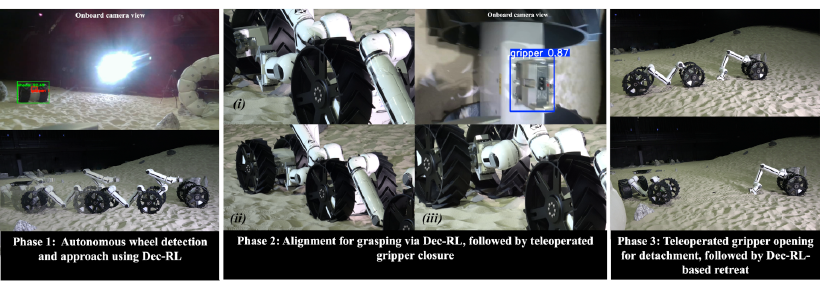}
    \caption{MoonBot reconfiguration in process: Phase 1: Dragon approaches with $\pi_{\text{wheel}}$ and $\pi_{\text{steer}}$, Phase 2: manipulator limb aligns with grasping fixture using $\pi_{\text{manip}}$, Phase 3: manipulator limb is detached using teleoperation to form Vehicle and Minimal modes.}
    \label{fig:reconfig_phases}
\end{figure*} 

\paragraph{Morphology-Specific Constraints}  
In \emph{Vehicle} mode, steering is restricted to joints 2 and 6 rotating about the \(z\)-axis in opposite directions, enabling in-place rotation of wheels while keeping other joints fixed to maintain limb rigidity and minimize oscillations. In \emph{Dragon} mode, the full 7-DoF articulation is exploited, including a central \(90^\circ\) joint to generate smooth triangular curvature between wheel bases, enhancing steering torque and stability. This modular policy reuse across diverse morphologies highlights the adaptability and robustness of our decentralized RL approach.


\section{Policy Composition and Coordination}

The proposed approach enables heterogeneous module policies to operate in unison without retraining, supporting three primary execution modes. The set of modules is denoted $\mathcal{M}$, each with a corresponding policy $\pi_m$ and local observation $o_t^m$. The global task $\mathcal{T}$ is decomposed by a mapping $\mathcal{C} : \mathcal{T} \to \mathcal{A}$, which assigns abstract commands to modules. At each timestep $t$, the coordination layer produces joint actions
\[
\mathbf{a}_t = \mathcal{F}\big(\{\pi_m(o_t^m)\}, s_t, \mathcal{C}(\mathcal{T})\big),
\]
where $s_t$ is the global state and $\mathcal{F}$ is the coordination function that merges local policy outputs with the task allocation.

\paragraph{Synchronous Parallel Execution}  
For a subset $\mathcal{M}_I \subseteq \mathcal{M}$ of identical modules, a shared policy $\pi_I$ receives a common observation $o_t^I$ and outputs identical actions for all modules in $\mathcal{M}_I$:
\[
a_t^{m} = \pi_I(o_t^{I}), \quad \forall m \in \mathcal{M}_I.
\]

\paragraph{Parallel Execution of Heterogeneous Policies}  
Modules of different types execute their respective policies concurrently:
\[
\mathbf{a}_t = \left( \pi_{\text{wheel}}(o_t^{\text{wheel}}), \pi_{\text{steer}}(o_t^{\text{steer}}), \ldots \right).
\]

\paragraph{Sequential Execution via Policy Gating}  
A binary gating variable $\alpha_t^m \in \{0,1\}$ controls whether module $m$ executes its policy output or an idle action $a_{\text{idle}}$:
\[
a_t^m = \alpha_t^m \pi_m(o_t^m) + (1-\alpha_t^m) a_{\text{idle}},
\]
with $\alpha_t^m = f_{\text{gate}}(s_t, \text{progress})$, where $f_{\text{gate}}$ determines activation based on global state and task progress.

In all cases, the coordination layer maintains a global task graph and synchronization map to ensure smooth transitions between execution modes. This enables adaptive reuse of local policies, scalability to new morphologies, and robust mission-aware control for structurally dynamic modular robots.

\section{Experimental Setup}




The proposed decentralized learning and policy composition approach was evaluated through hardware experiments in the lab and in a lunar-analog sandy field at JAXA’s Advanced Facility for Space Exploration~\cite{jaxafield}. In the lab, we tested only the manipulation policy on the 7-DoF limb for precision grasping (Fig.~\ref{fig:policy_performance}(a)). The field test aimed to demonstrate autonomous modular reconfiguration under granular terrain, uneven lighting, and obstacles. We used the \emph{Dragon} configuration (two wheel–limb modules) to validate integration of independently learned policies for locomotion, steering, and manipulation, where the \emph{Dragon} morphology must detect, navigate to, and physically connect a separate, unanchored wheel module located within the test field. No prior global position is provided for the target; instead, the robot relies solely on onboard YOLO-based perception \cite{redmon_yolo_2016} and the learned modular control policies to accomplish the operations. In application, Yolo based perception can be easily replaced by any other algorithm providing relative target coordinates. This setup provides a realistic platform to test the robustness, adaptability, and fault tolerance of decentralized learning and modular composition, emulating mission tasks like in-situ hardware reconfiguration, asset recovery, and adaptive morphological expansion in extravehicular environments. 

The reconfiguration process unfolds sequentially in three phases, as presented in Fig. \ref{fig:reconfig_phases}. Initially, the onboard perception systems visually localize the target wheel module, converting its relative pose into goal coordinates for the locomotion RL policy control stack. The wheel locomotion policy \(\pi_{\text{wheel}}\) is deployed synchronously on both the front and rear wheel modules to ensure stable forward motion. Concurrently, the bridge-mounted 7-DoF limb executes the steering policy \(\pi_{\text{steer}}\), dynamically adjusting the heading and trajectory based on visual sensor feedback, thereby accommodating the lunar-like sandy, i.e., slippery, test terrain. Upon reaching a proximity threshold (approximately \SI{0.3}{\meter}), a high-level supervisory MDP triggers an acknowledgment signal \(\alpha_t = 1\), which simultaneously deactivates the locomotion and steering policies and activates the manipulation policy \(\pi_{\text{manip}}\). The Dragon's front 7-DoF manipulator limb then performs alignment using this manipulator policy. At this stage, the policy operates in an open-loop control mode once the \emph{Dragon} halts. Based on the coordinates of the wheel grapple fixture, the policy generates a sequence of joint states while considering the current joint state of the limb. Final adjustment and grasping of the target module’s interface are carried out through human-in-the-loop teleoperation, as the fixture requires precise overlap, while the manipulator policy is designed only to achieve alignment within a \SI{0.10}{m} radius. Once successful mechanical attachment is confirmed through human-in-the-loop feedback, the manipulator is detached from the \emph{Dragon}'s front wheel module also via teleoperation, and the system reconfigures into \emph{Vehicle} and \emph{Minimal} modes. The resulting morphologies can then be driven using their respective Dec-RL policies, teleoperation, or classical controllers, depending on missions requirements.

Ground truth target and robot positions are recorded using OptiTrack motion capture (MoCap) and visual tracking systems to facilitate the quantitative evaluation of navigation accuracy and manipulation success. 

\section{Results and Discussion}
Unless noted otherwise, all curves report mean $\pm$ \emph{s.d.} over 10 random seeds. Our simulation environments do not explicitly model regolith particles but incorporate domain randomization over terrain friction, sensor noise, and delays to support robust policy learning. These policies were transferred directly to hardware and tested on granular lunar-like terrain at JAXA’s Space Exploration Field without fine-tuning, enabling evaluation of zero-shot generalization across modules and configurations.

\subsection{Policy Performance during Learning}
\begin{figure}[t]
    \centering
    \includegraphics[width=0.9\linewidth]{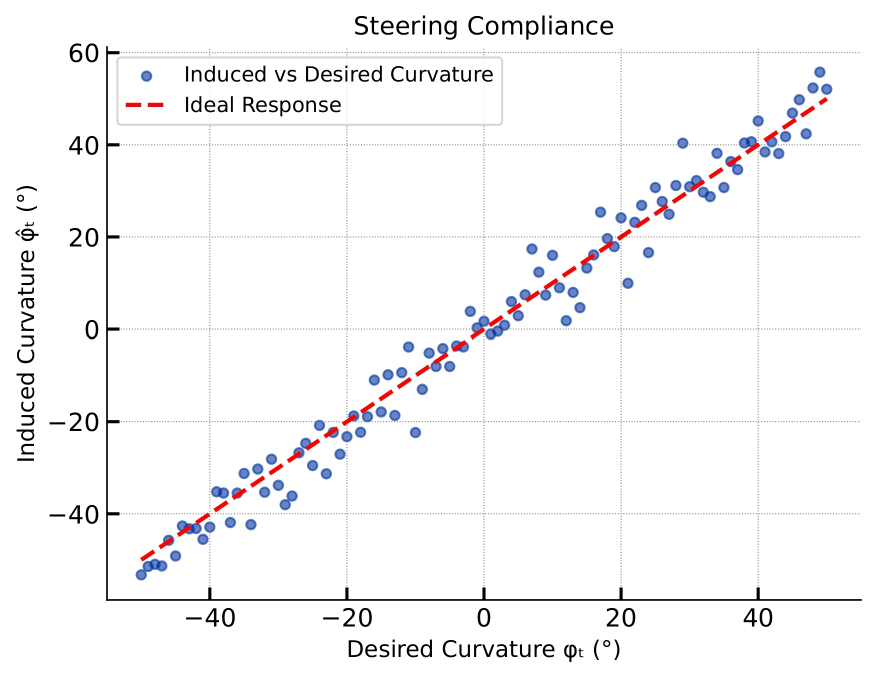}
    \caption{Steering compliance of the 7-DoF limb Policy: induced vs. desired curvature.}
    \label{fig:steering_compliance}
\end{figure}

Fig.~\ref{fig:steering_compliance} illustrates the relationship between desired ($\phi_t$) and induced ($\hat{\phi}_t$) steering curvatures for the 7-DoF limb executing the learned steering policy in simulation. The system achieved a Pearson correlation coefficient ($r$) of 0.9895 and a coefficient of determination ($R^2$) of 0.9758, confirming a strong linear correspondence across the full curvature range of $-50^\circ$ to $+50^\circ$.

The mean absolute error (MAE) and root-mean-square error (RMSE) were \SI{3.63}{\degree} and \SI{4.54}{\degree}, respectively. A total of \SI{78.2}{\percent} of commands were executed within \SI{5}{\degree} of the target, with \SI{30.7}{\percent} within \SI{2}{\degree}. The tight clustering of data points along the ideal response line indicates that the steering limb, positioned between the two wheel modules in Dragon mode, consistently converts target curvatures into accurate steering motions while sustaining performance across the entire operational range.

\begin{figure}[t]
    \centering
    \includegraphics[width=\linewidth]{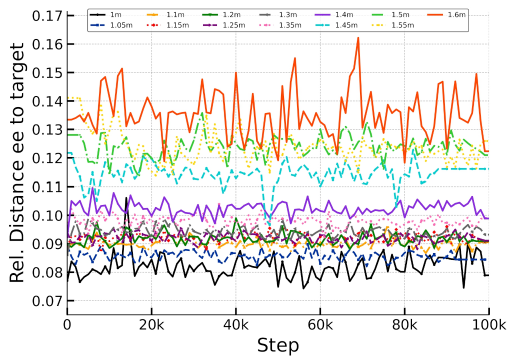}
    \caption{Relative distance between end-effector and target across varying initial distances.}
    \label{fig:ee_target}
\end{figure}

The learning dynamics of the 7-DoF manipulation policy across 13 target distances (\SI{1.0}{m} - \SI{1.6}{m}) is shown in Fig.~\ref{fig:ee_target}. All series reached improvement stagnation within an average of 1466 steps, with \SI{84.6}{\percent} exhibiting low variance after improvement plateau. A practical accuracy threshold of \SI{0.10}{\meter} relative distance was met in \SI{53.8}{\percent} of cases (7/13), primarily for targets within the \SI{1.0}{m} - \SI{1.3}{m} range, with an average convergence time of 1289 steps. Final positioning error increased with distance, ranging from 0.0832 (\SI{1}{m}) to 0.1369 (\SI{1.6}{m}), showing a strong positive correlation ($r = 0.937$) between target range and residual error. These results indicate robust convergence behavior while highlighting distance-dependent accuracy scaling.

\begin{figure}[t]
    \centering
    \includegraphics[width=0.9\linewidth]{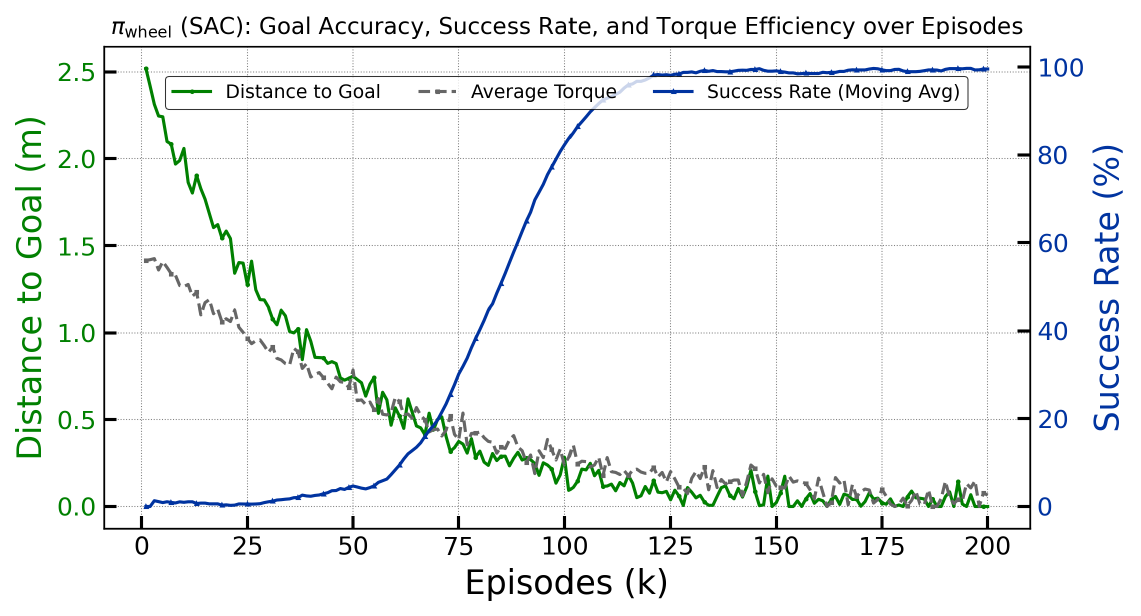}
    \caption{Wheel locomotion policy performance: distance-to-goal, success rate, and torque usage over training episodes.}
    \label{fig:wheel_policy}
\end{figure}

The wheel locomotion policy $\pi_{\text{wheel}}$, trained using SAC, exhibited rapid improvement and high efficiency (Fig.~\ref{fig:wheel_policy}). Goal-reaching performance improved from an initial \SI{2.520}{m} to \SI{0.000}{m} final distance (showing \SI{100}{\percent} improvement), with the success rate increasing from \SI{0}{\percent} to \SI{99.6}{\percent}. A \SI{50}{\percent} reduction in distance was achieved by episode~27, sub-\SI{0.5}{m} accuracy by episode~72, and $>\SI{80}{\percent}$ success rate by episode~99. Average torque demand decreased by \SI{95.4}{\percent} (from \SI{1.414}{Nm} to \SI{0.065}{Nm}), indicating substantial energy efficiency gains. Training stability was evidenced by a success rate coefficient of variation of only 0.003 over the final \SI{20}{\percent} episodes, reflecting consistent high-performance. The observed learning efficiency of 0.50 success-percentage points per episode, combined with perfect final distance accuracy, confirms the effectiveness of the SAC algorithm for the wheel navigation task.

\subsection{Performance Evaluation on Hardware}
\begin{figure}
    \centering
    \begin{minipage}[b]{0.45\linewidth}
        \centering
        \includegraphics[width=\linewidth]{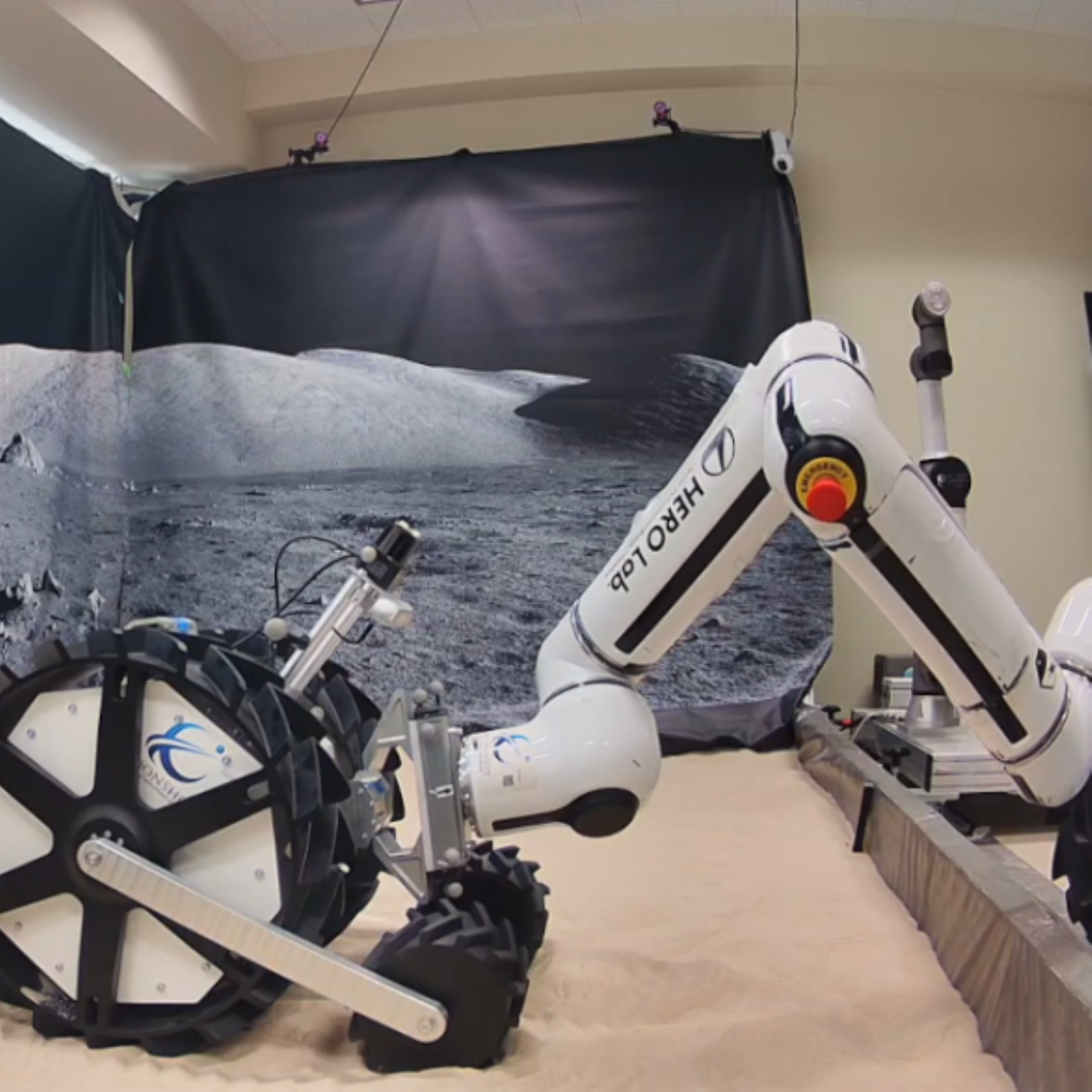}
        \vspace{2pt}
        
        {\footnotesize (a)}
        \label{fig:in_lab_setup}
    \end{minipage}%
    \hfill
    \begin{minipage}[b]{0.52\linewidth}
        \centering
        \includegraphics[width=\linewidth]{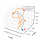}
        \vspace{2pt}
        
        {\footnotesize (b)}
        \label{fig:arm_policy_traj}
    \end{minipage}
    \caption{In Lab Testing: (a) Hardware Setup, (b) RL Policy Trajectory towards wheel}
    \label{fig:policy_performance}
\end{figure}

Algorithms were selected as standard, reliable choices to demonstrate multi-modal modular deployment; benchmarking alternatives is left to future work. While conventional model‑based or IK‑based planners can also achieve task completion, they require separate design and tuning for each configuration. In contrast, our multi‑modal modular RL framework demonstrates that reusable policies can be composed dynamically across morphologies, which is the central scope of this work.
The Manipulator RL policy was evaluated in a laboratory sand-bed environment (Fig. \ref{fig:policy_performance}(a)) using the 7-DoF manipulator limb for precision grappling tasks. While YOLO-based target coordinate extraction and autonomous approach are possible since the policy only requires the target’s coordinates, precise evaluation and analysis of backlash-induced wobble effects required integration of a MoCap system. This mechanical issue cause joints to move freely to some extent, and hence the MoCap system provided real-time end-effector pose measurements with \SI{1}{\milli\meter} precision. The experimental comparison in Fig. \ref{fig:policy_performance}(b) highlights two distinct control modes. In the first mode, the wheel grapple fixture coordinates were provided to the RL policy only at initialization. The generated joint states were then executed sequentially directly on hardware forming an open-loop control. This resulted in a smooth trajectory (dashed blue line) with an efficient \SI{1.55}{\meter} path length and high positioning accuracy of \SI{5.51}{\centi\meter} from the target. However, mechanical backlash caused a small offset between the computed and actual end-effector positions. In the second mode, coordinate feedback was provided continuously to the policy in real time. Due to backlash, the computed end-effector position was offset, and feeding these erroneous coordinates back to the system caused the policy to repeatedly attempt compensation. This led to a series of over-corrections, producing persistent oscillations with a \SI{3.9}{\times} longer path length (\SI{6.10}{\meter}), \SI{43}{\percent} higher jerk values (indicating reduced smoothness), and a degraded final positioning accuracy of \SI{19.20}{\centi\meter}. These results show that, in this setup, real-time feedback compensation can amplify mechanical nonlinearities rather than correct them, leading to the characteristic “endless wobbling” that reduces task performance. Due to these reasons, during the autonomous reconfiguration demonstrated in Fig. \ref{fig:reconfig_phases}, open-loop mode was used with human-in-the-loop feedback through tele-operation for final adjustment to offset and grasping.

\begin{figure}[b]
    \centering
    \begin{minipage}{0.48\linewidth}
        \includegraphics[width=\linewidth]{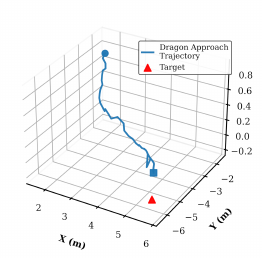}
    \end{minipage}\hfill
    \begin{minipage}{0.48\linewidth}
        \includegraphics[width=\linewidth]{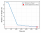}
    \end{minipage}
    \caption{Sim2Real Implementation Dragon Mode: (Left) Trajectory of Dragon Locomotion in 3D space, (right) Relative distance to target during Locomotion}
    \label{fig:dragon_approach_traj}
\end{figure}

The experimental trajectory (Fig. \ref{fig:dragon_approach_traj}) illustrates the real-time performance of the coordinated wheel and steering policies as the \emph{Dragon} moves from its initial position at (\SI{1.527}{\meter}, \SI{-1.515}{\meter}, \SI{0.687}{\meter}) to the target position at (\SI{5.819}{\meter}, \SI{-6.503}{\meter}, \SI{-0.024}{\meter}) over a total duration of \SI{319.19}{s}. The motion profile reveals a characteristic three-phase behavior. In the initial exploration phase, the robot exhibits high positional variance (\SI{2.79}{m^2}) as the YOLO-based visual feedback system acquires target lock and the policies adapt to real-world dynamics. This is followed by a refinement phase, where stabilization improves dramatically with a \SI{98.9}{\percent} reduction in variance (down to \SI{0.032}{\meter\squared}). In the final phase, precise terminal guidance is achieved, with exceptionally low variance (\SI{0.006}{\meter\squared}) and a final approach accuracy of \SI{0.338}{\meter}. Over the course of the experiment, the robot travels \SI{8.092}{\meter}, achieving an efficient \SI{81.3}{percent} path efficiency ratio relative to the optimal straight-line distance of \SI{6.582}{\meter}. The motion remains smooth, as indicated by a low velocity standard deviation of \SI{0.030}{\meter\per\second} and controlled path curvature averaging \SI{0.498}{\radian\per\meter}, with a minimal approach angle deviation of \SI{13.2}{\degree} from the optimal direct trajectory. Overall, this real-time hardware validation demonstrates the framework’s ability to achieve robust target convergence through coordinated policy execution, reducing the initial \SI{6.619}{\meter} target distance by \SI{94.9}{\percent} while maintaining a deliberate, precision-oriented motion at an average velocity of \SI{0.025}{\meter\per\second}, well-suited for safe manipulation in unstructured environments.

\begin{figure}[t]
    \centering
    \includegraphics[width=\linewidth]{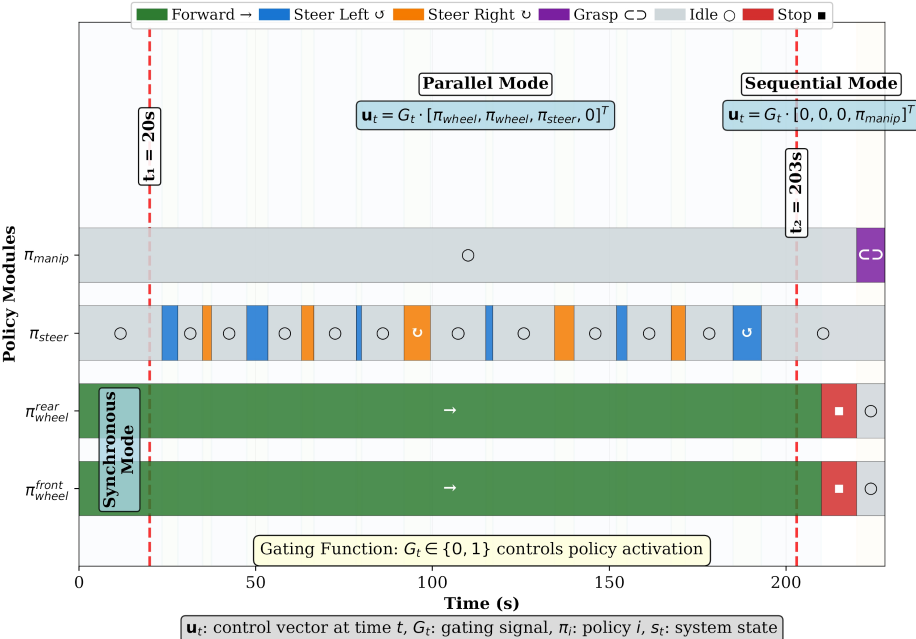}
    \caption{Unified Gantt-style execution timeline for synchronous, parallel, and sequential policy modes, coordinated via a high-level gating function.}
    \label{fig:policy_coordination}
\end{figure}

The integrated coordination of four policies during the second trial, front and rear wheel locomotion, steering, and manipulation, over a \SI{228}{\second} mission is summarized in Figure~\ref{fig:policy_coordination}. 

Synchronous execution of both wheel policies accounted for \SI{70.9}{\percent} of the timeline, enabling stable forward motion. Parallel execution, integrating steering with locomotion, comprised \SI{21.0}{\percent} of the mission and included 96 steering events (50 left turns, 46 right turns). Sequential execution, activating only the manipulation policy, accounted for \SI{3.7}{\percent} of the timeline, during which grasp alignment operation were performed. The approach maintained a \SI{100}{\percent} activity ratio with only 24 smooth state transitions, averaging \SI{9.5}{s} apart, demonstrating robust arbitration without idle states or control conflicts. Overall, the results demonstrate that independently trained, module-specific policies, when composed via a lightweight coordination layer, can reliably achieve execution of complex multi-stage tasks in dynamically reconfigurable robots. 


\section{Conclusions}
This work introduced a decentralized RL approach for modular lunar robots, validated in full-scale lab trials and at JAXA’s Space Exploration Test Field. The method improved steering precision, stabilized manipulation, reduced torque use, and enabled coordinated operation, demonstrating scalability and robustness. Autonomous locomotion, steering, and pre-grasp alignment were achieved; final grasping and disconnection were teleoperated due to safety and hardware limitations, such as gear backlash. Validation focused on morphologies (Vehicle and Dragon); that could traverse on loose sand and minimize weight-related sinkage, allowing for self-reconfiguration.
We will broaden coverage with lighter hardware and contact-aware control. Future work targets higher sample efficiency across tasks, policy reuse across more morphologies, and greater robustness toward a scalable, adaptable control solution for long-duration, reconfigurable space operations.

\section*{Acknowledgement}
The authors would like to thank Jakob Madestam and Marcus Dyhr for their valuable support in conducting the experiments.

\bibliographystyle{./IEEEtran}
\bibliography{reference.bib}

\end{document}